\PassOptionsToPackage{dvipsnames}{xcolor}
\PassOptionsToPackage{colorlinks=true, linkcolor=blue, citecolor=blue}{hyperref}
\documentclass[11pt, a4paper, logo, copyright, nonumbering]{openreasonerzero}

\usepackage[utf8]{inputenc} 
\usepackage[T1]{fontenc}    
\usepackage{hyperref}       
\usepackage{url}            
\usepackage{booktabs}       
\usepackage{amsfonts}       
\usepackage{amsmath}        
\usepackage{nicefrac}       
\usepackage{microtype}      
\usepackage{xcolor}         

\usepackage{graphicx}
\usepackage{enumitem}
\usepackage{multirow}  
\usepackage[capitalise]{cleveref}
\usepackage[normalem]{ulem}
\usepackage{makecell}
\usepackage[square,numbers]{natbib}
\usepackage{wrapfig}
\usepackage{siunitx}
\usepackage{tikz}  
\usepackage{caption}

\definecolor{AliceBlue}{rgb}{0.94, 0.97, 1.0}

\fancypagestyle{firstpage}{
  \fancyhf{} 
  \fancyhead[L]{%
    \raisebox{-0.4\height}{\includegraphics[height=40pt]{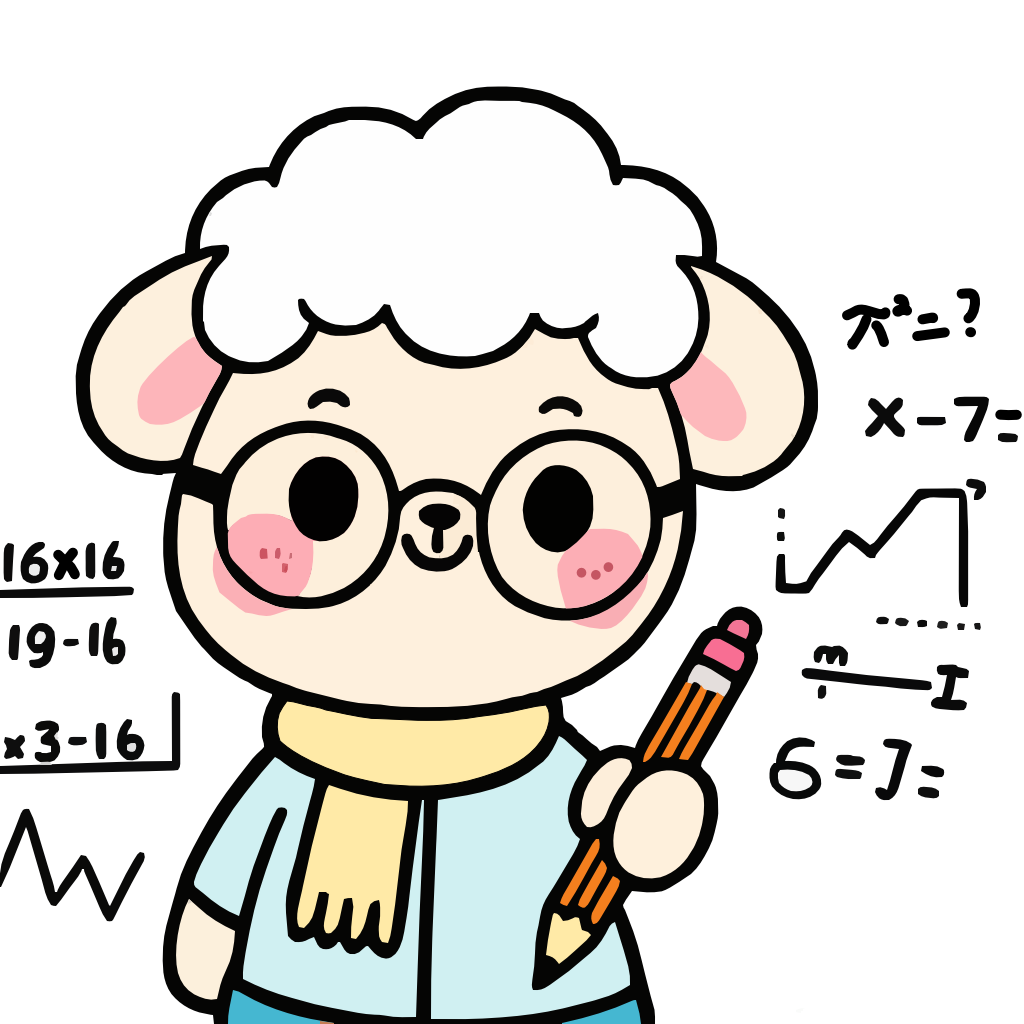}} \hspace{0.5em}
    {\small \textit{\the\month-\the\day-\the\year}}%
  }
  \fancyfoot[L]{\small * Core contribution \\ $\dagger$~Corresponding authors: zhaoliang02@stepfun.com, vpatel36@jhu.edu}
  
}

\title{
  Open Vision Reasoner: Transferring Linguistic Cognitive Behavior for Visual Reasoning
}

\author{

Yana Wei$^{1*}$, Liang Zhao$^{2*\dagger}$, Jianjian Sun$^{2*}$, Kangheng Lin$^{3}$, Jisheng Yin$^{4}$,
Jingcheng Hu$^{5}$, Yinmin Zhang$^{2}$, 
En Yu$^{6}$,
Haoran Lv$^{2}$, Zejia Weng$^{2}$, Jia Wang$^{2}$, Chunrui Han$^{2}$, Yuang Peng$^{5}$, 
Qi Han$^{2}$, Zheng Ge$^{2}$,
Xiangyu Zhang$^{2}$, Daxin Jiang$^{2}$, 
Vishal M. Patel$^{1\dagger}$\\
\small{$^1$Johns Hopkins University \quad $^2$StepFun \quad
$^3$BUPT \quad $^4$UCAS \quad $^5$THU\quad $^6$HUST\\}

}

\begin{document}

\begin{abstract}

The remarkable reasoning capability of large language models (LLMs) stems from cognitive behaviors that emerge through reinforcement with verifiable rewards.
This work investigates how to transfer this principle to Multimodal LLMs (MLLMs) to unlock advanced visual reasoning. We introduce a two-stage paradigm built on Qwen2.5-VL-7B: a massive linguistic cold-start fine-tuning, 
followed by multimodal reinforcement learning (RL) spanning nearly 1,000 steps—surpassing all previous open-source efforts in scale.
This pioneering work reveals three fundamental insights:
1) Behavior transfer emerges surprisingly early in cold start due to linguistic mental imagery.
2) Cold start broadly memorizes visual behaviors, while RL critically discerns and scales up effective patterns.
3) Transfer strategically favors high-utility behaviors such as visual reflection.
Our resulting model, Open-Vision-Reasoner (OVR), achieves state-of-the-art performance on a suite of reasoning benchmarks, including 95.3\% on MATH500, 51.8\% on MathVision and 54.6\% on MathVerse. We release our model, data, and training dynamics to catalyze the development of more capable, behavior-aligned multimodal reasoners.

\end{abstract}
\maketitle
\thispagestyle{firstpage}  

\begin{figure*}[htbp]
    \vspace{-3mm}
    \centering
    \includegraphics[width=\textwidth]{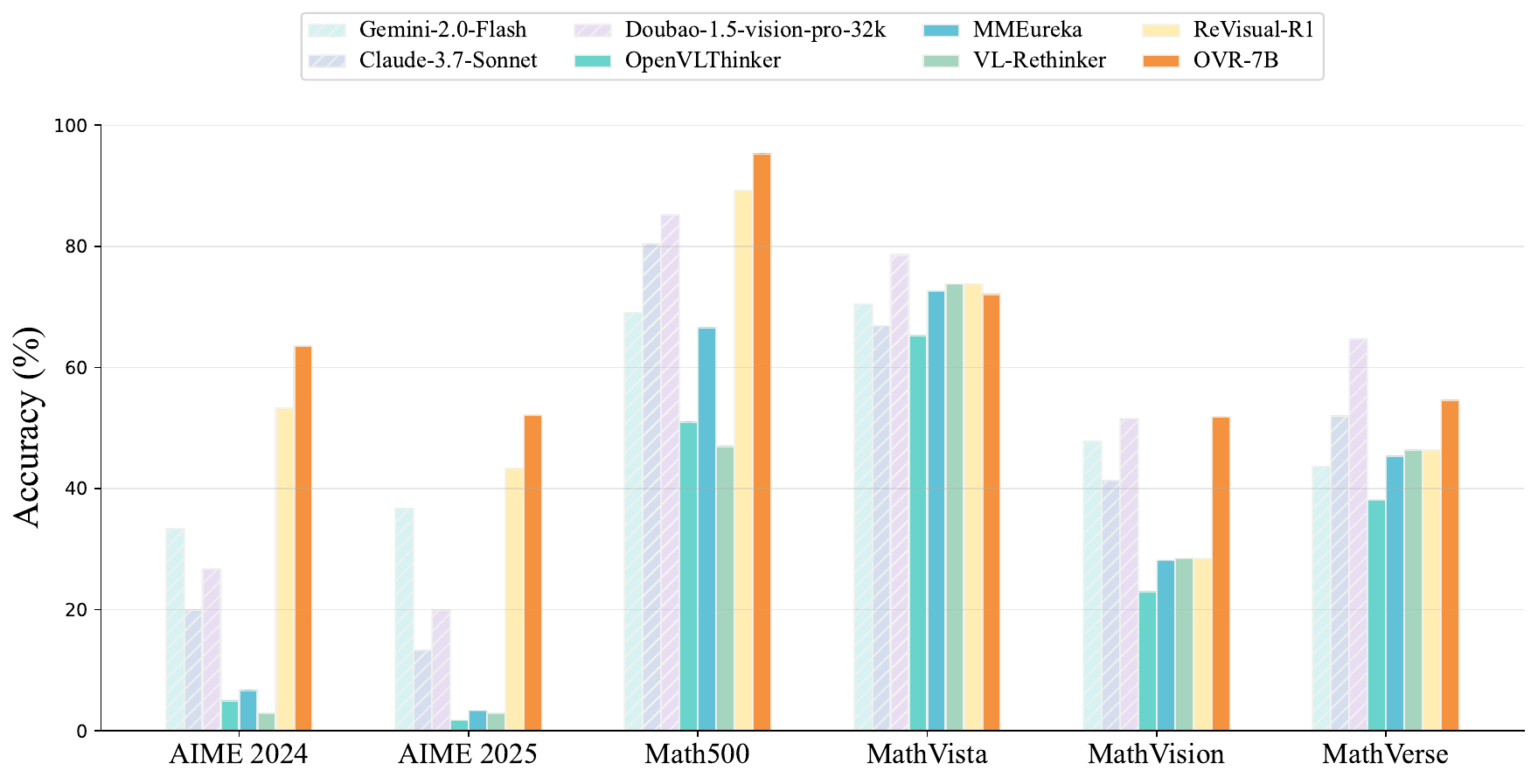}
\caption{
\textbf{Performance comparison} with state-of-the-art models on both textual (AIME 2024, AIME 2025~\citep{balunovic_srimatharena_2025}, MATH500~\citep{hendrycksmath2021}) and multimodal (MathVista~\citep{lu2023mathvista}, MathVision~\citep{wang2024measuring}, MathVerse~\citep{zhang2024mathversedoesmultimodalllm}) math reasoning benchmarks. \textbf{Open Vision Reasoner (OVR)} demonstrates superior results among open-source models and performs competitively with commercial counterparts.
}
    \vspace{-10mm}
    \label{fig:per-overview}
\end{figure*}

\section{Introduction}
\label{sec:intro}

\begin{quote}
\textit{``The eye sees only what the mind is prepared to comprehend.''} --- Robertson Davies
\end{quote}

Shifting Reinforcement Learning from Human Feedback (RLHF)~\citep{ouyang2022training} to Reinforcement Learning from Verifiable Reward (RLVR)~\citep{dsr1_cite,kimi1p5_cite} has endowed LLMs~\citep{openaio1_cite,dsr1_cite} with unexpectedly powerful reasoning across mathematics, code, and general problem-solving. At its core, verifiable reward—where correctness is determined by objective, often rule-based criteria—is inherently less susceptible to "reward hacking"~\citep{rewardhack,yu2025unhackable} than a learned reward model. This robustness proves instrumental in large-scale RL, enabling the internalization and activation of what recent studies~\citep{yu2024rlhf,gandhi2025cognitive,zhao2025echo} term \textbf{cognitive behaviors}—patterns like backtracking and subgoal decomposition that are empirically crucial for advanced reasoning.

The multimodal domain, inherently grounded in verifiable visual facts~\citep{chen2023shikra,yu2025perception}, is uniquely suited for this paradigm. Yet, early multimodal RL efforts paradoxically adopted RLHF, relying on learned reward models to approximate objective correctness~\citep{wang2024mdpo,zhu2025perpo,zhu2024self}. Inspired by the success of RLVR in language models, recent efforts have started exploring rule-based rewards in the multimodal setting. 
Perception-R1~\citep{yu2025perception} incorporates supervisions such as IoU and Euclidean distance to enhance the perceptual alignment of MLLMs, while works such as R1-OneVision~\citep{yang2025r1} and VLAA-Thinking~\citep{chen2025sft} construct behavior-rich visual reasoning trajectories through complex pipelines including iterative distillation and synthesizing. Recently, ReVisual-R1~\citep{revisualr1} adopts a effective language-only cold start as a foundation for visual reasoning.


Despite this encouraging progress, these approaches still leave a foundational question unanswered:
\textbf{How can linguistic cognitive behaviors transfer to MLLMs for advanced visual reasoning?} 
To address this, we build upon the "RL with a cold start" paradigm~\citep{dsr1_cite} by conducting large-scale training on Qwen2.5-VL-7B~\citep{bai2025qwen2dot5}, establishing it as a powerful testbed to systematically analyze how such behaviors emerge and scale in the multimodal domain.

To this end, we introduce a robust two-stage methodology designed to first instill linguistic cognitive patterns and then activate them for visual reasoning. Our process begins with a large-scale cold start, fine-tuning Qwen2.5-VL-7B on over \textit{2 million} examples to build a strong foundation. This is followed by a prolonged reinforcement learning phase under the \textit{Open-Reasoner-Zero}~\citep{hu2025open} framework, leveraging over \textit{0.3 million} mixed-modality examples. 
To the best of our knowledge, this represents the largest open-source RL practice on this model. The resulting model, \textbf{Open-Vision-Reasoner} (OVR), validates our approach by achieving strong performance across both language and multimodal benchmarks. As shown in \cref{fig:per-overview}, it achieves \textbf{63.5\%} on AIME2024 and \textbf{95.3\%} on MATH500 for math reasoning, as well as \textbf{51.8\%} on MathVision and \textbf{54.6\%} on MathVerse for visual reasoning.


To further trace the transfer and evolution of cognitive patterns throughout training, we develop a in-depth \textbf{visual cognitive behavior analysis}. Three central insights are worth highlighting:
(1) Behavior transfer emerges remarkably early in cold start, driven by linguistic patterns encoding \textit{mental imagery}~\citep{ms1,ms2} as illustrated in~\cref{fig:r1-lang-case}.
(2) Cold start broadly \textit{memorizes} diverse visual cognitive behaviors, while RL critically \textit{discerns} and scales up effective patterns.
(3) Transfer follows a \textit{strategic} path, favoring behaviors with high utility such as visual reflection.
These findings deepen the understanding on visual intelligence scaffolded by linguistic reasoning~\citep{virgo}.

We further examine how this paradigm impacts a foundational capability of MLLMs—\textbf{visual perception}. While linguistic cold start introduces perceptual degradation, our study shows that multimodal RL can effectively \textit{recover} this loss. However, we also observe the \textit{limited scalability} of RL when focused solely on perceptual tasks, as reward signals increase without corresponding growth in reasoning complexity (e.g., token length). 
This limitation motivates a more deliberate integration of diverse, primitive visual cognitive behaviors. Such efforts represent a promising direction toward unlocking the potential of more advanced RL frameworks—multi-turn or even agentic RL built upon visual manipulation and imagination.

In summary, this paper advances the field through the following \textbf{three key contributions}:
\begin{itemize}[leftmargin=1.2em]
\item We construct a two-stage training pipeline consisting of a linguistic cold start followed by large-scale multimodal RL, enabling effective transfer of cognitive behaviors in MLLMs.
\item Our \textit{Open Vision Reasoner}, the largest open-source RL practice on Qwen2.5-VL-7B, achieves superior performance on both linguistic and multimodal reasoning benchmarks.
\item We conduct an in-depth analysis of visual cognitive behaviors in OVR and provide valuable insights into their transfer and evolution across training stages.
\end{itemize}

\begin{table}[t]
\centering
\small
\renewcommand{\arraystretch}{1.2}
\caption{\textbf{Visual Cognitive Behaviors and Linguistic Counterparts.} We define four key visual cognitive behaviors, providing formal definitions, illustrative examples, and their corresponding linguistic counterparts.}
\resizebox{\textwidth}{!}{%
\begin{tabular}{
>{\raggedright\arraybackslash}p{3.2cm} 
>{\raggedright\arraybackslash}p{3.2cm} 
>{\raggedright\arraybackslash}p{6.5cm} 
>{\raggedright\arraybackslash}p{3cm}}
\toprule
\textbf{Visual Behavior} & \textbf{Example} & \textbf{Definition} & \textbf{Linguistic Counterpart} \\
\midrule
\textbf{Visual Reflection} & ``Let me see the image again.'' & The model explicitly revisits the image after identifying a potential mistake or inconsistency in its reasoning, indicating an effort to correct course. & Backtracking \\
\addlinespace
\textbf{Divide-and-Conquer} & ``Let’s first look at the numbers on the left.'' & The model breaks down a complex visual problem into sub-components or regions, each addressed sequentially to reach the final answer. & Subgoal Setting \\
\addlinespace
\textbf{Visual Verification} & ``I will now verify this against the image.'' & The model confirms that its intermediate conclusions are visually grounded by cross-referencing with the image before proceeding. & Verification \\
\addlinespace
\textbf{Goal-driven Visual Tracing} & ``To get this answer, I need to find an object that...'' & The model starts from a desired visual conclusion and reasons backwards to identify relevant image evidence that supports it. & Backward Chaining \\
\bottomrule
\end{tabular}
}
\label{tab:visual-cog-def}
\end{table}
\section{Cognitive Behavior Preliminaries}
\label{sec:pre}

Recent studies have highlighted that the emergence of robust reasoning in LLMs is closely tied to the acquisition of certain \textit{cognitive behaviors}~\citep{gandhi2025cognitive}. These behaviors reflect structured internal reasoning patterns akin to human problem-solving. Key examples include:
(1) Backtracking — revising a previously chosen strategy upon identifying inconsistencies (e.g., “This approach won’t work because…”),
(2) Verification — checking intermediate steps or partial results,
(3) Subgoal Setting — breaking down complex problems into manageable components (e.g., “First, we need to…”),
and (4) Backward Chaining — reasoning from the desired outcome back to required inputs (e.g., “To get 75, we need a number divisible by…”).
The four patterns form a kind of textual inner monologue that emerges naturally in language-based reasoning.

\begin{figure}[t]
    \centering
    \includegraphics[width=\textwidth]{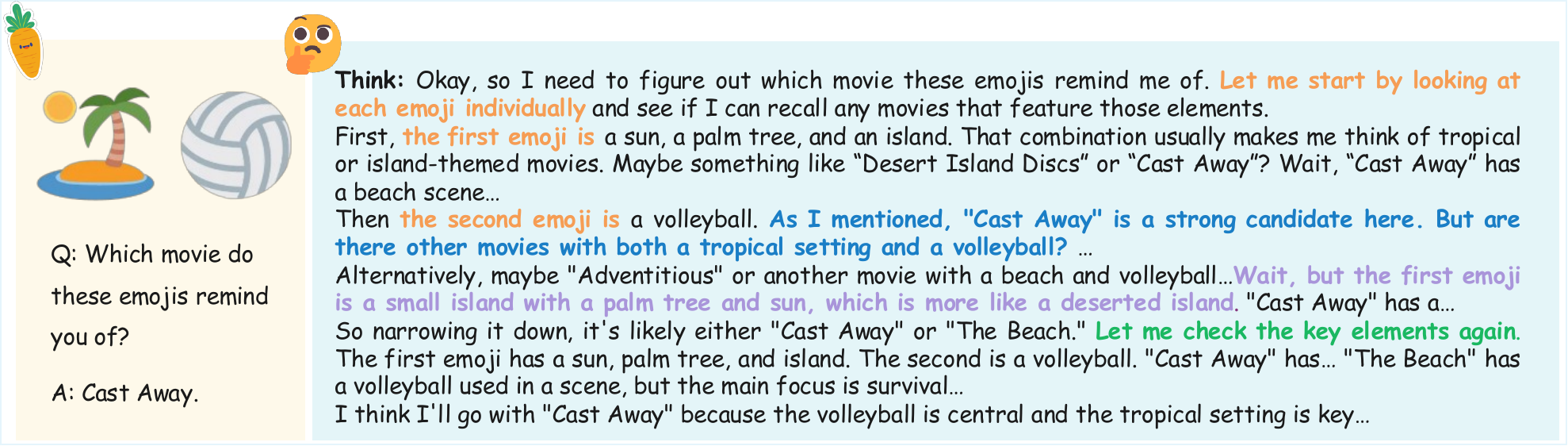}
    \caption{\textbf{Multiple Cognitive Behaviors in a Single Response.} This case shows triggered visual-specific cognitive behaviors like \textcolor[HTML]{D55E00}{\textbf{visual divide-and-conquer}}, \textcolor[HTML]{377129}{\textbf{reflection}}, \textcolor[HTML]{8975bd}{\textbf{goal-driven visual tracing}}, along with the linguistic behavior \textcolor[HTML]{0072B2}{\textbf{backtracking}}.}
    \label{fig:case}
\end{figure}

Based on this, we investigate the \textit{transfer} of cognitive behaviors from language to vision. 
We define the visual extensions of the aforementioned behaviors—visual reflection, divide-and-conquer, visual verification, and goal-driven visual tracing. Their formal definitions, examples, and corresponding linguistic counterparts are provided in~\cref{tab:visual-cog-def}, while~\cref{fig:case} presents a multimodal example encompassing both linguistic and visual cognitive behaviors.
In the following sections, we present a simple yet effective MLLM training pipeline comprising a linguistic cold start followed by multimodal RL (\cref{sec:methods}), and systematically analyze the transfer and scaling of these visual cognitive behaviors (\cref{sec:exp_visual_behavior}).
\section{Open-Vision-Reasoner}
\label{sec:methods}

In this section, we introduce \textbf{Open-Vision-Reasoner} (OVR), a strong multimodal reasoning model build from Qwen2.5-VL-7B~\cite{qwen2p5_cite}, from perspectives of training pipeline (\cref{sec:pipeline}), RL algorithm (\cref{sec:rl_alg}) and data construction (\cref{sec:data_cur}).

\subsection{Training Pipeline}
\label{sec:pipeline}

To facilitate efficient cognitive development and cross-modal generalization, we employ the popular "RL with a cold start" paradigm~\citep{dsr1_cite} with two sequential training stages:

\begin{itemize}[leftmargin=*]
    \item \textbf{Stage 1: Linguistic Cold Start.} The LLM module is supervised fine-tuned on language-only reasoning datasets distilled from DeepSeek-R1~\citep{dsr1_cite}, establishing core cognitive behaviors such as backtracking and subgoal decomposition within a purely linguistic setting.
    \item \textbf{Stage 2: Multimodal RL.} We apply reinforcement learning with Open-Reasoner-Zero~\citep{hu2025open} setting on both text and multimodal tasks using verifiable match rewards. This promotes reasoning generalization and aligns previously learned cognitive patterns with visual contexts, enabling effective cross-modal transfer.
\end{itemize}

\subsection{RL Algorithm}
\label{sec:rl_alg}

For the RL stage of our training pipeline, we adopt a lightweight Proximal Policy Optimization (PPO)~\citep{ppo_cite} with Generalized Advantage Estimation (GAE)~\citep{gae_cite}, following the policy and reward design used in \textit{Open-Reasoner-Zero}~\citep{hu2025open}. We detail the RL for multimodal tasks below:

\paragraph{Proximal Policy Optimization}
For each input, consisting of an image $I$ and a textual prompt $q$, the policy network $\pi_\theta$ generates $n$ responses $\{o_1, \ldots, o_n\}$. Each response $o_i$ is a trajectory $\tau_i = (s_0^{(i)}, a_0^{(i)}, \ldots, s_{T_i-1}^{(i)}, a_{T_i-1}^{(i)})$ of length $T_i$. The state $s_t^{(i)}$ includes $q$ (and potentially encoded $I$ features) and previously generated tokens; $a_t^{(i)}$ is the token generated at step $t$. A reward $r_t^{(i)}$ is computed at each timestep $t$ of trajectory $\tau_i$.

We use GAE to balance bias and variance in advantage estimation. The advantage $\hat{A}_t^{(i)}$ for state-action pair $(s_t^{(i)}, a_t^{(i)})$ in trajectory $\tau_i$ is:
\begin{equation}
\hat{A}_t = \sum_{l=0}^{T-t-1} (\gamma \lambda)^l \delta_{t+l}, \quad \text{where } \delta_{t'} = r_{t'} + \gamma V_\phi(s_{t'+1}) - V_\phi(s_{t'}).
\end{equation}
$V_\phi$ is the value function, $\gamma, \lambda$ are discount and GAE factors, and $V_\phi(s_T)=0$ for terminal states. $\pi_\theta$ is updated by maximizing $\mathcal{J}_{\text{PPO}}(\theta)$ using experiences $(s_t, a_t, \hat{A}_t)$ sampled under an older policy $\pi_{\text{old}}$:
\begin{equation} \label{eq:ppo_ultra_concise}
\mathcal{J}_{\text{PPO}}(\theta) = \hat{\mathbb{E}}_{\pi_{\text{old}}} \left[ \min \left( \rho_t(\theta) \hat{A}_t, \text{clip} \left( \rho_t(\theta), 1 - \epsilon, 1 + \epsilon \right) \hat{A}_t \right) \right].
\end{equation}
Here, $\rho_t(\theta) = \frac{\pi_\theta(a_t | s_t)}{\pi_{\text{old}}(a_t | s_t)}$ and $\epsilon$ is a clipping parameter (e.g., $0.2$). $\hat{\mathbb{E}}_{\pi_{\text{old}}}$ denotes the empirical average over samples from $\pi_{\text{old}}$.
$V_\phi$ is trained by minimizing $\mathcal{J}_{\text{value}}(\phi)$ on samples from $\pi_{\text{old}}$, using the empirical discounted returns $R_t = \sum_{k=0}^{T-t-1} \gamma^k r_{t+k}$:
\begin{equation} \label{eq:value_ultra_concise}
\mathcal{J}_{\text{value}}(\phi) = \hat{\mathbb{E}}_{\pi_{\text{old}}} \left[ \left( V_\phi(s_t) - R_t \right)^2 \right].
\end{equation}

\paragraph{Reward Function.} 
We adopt the minimalist rule-based reward design, which evaluates only the correctness of model outputs while ignoring formatting or stylistic preferences. 
Specifically, we extract the predicted answer encapsulated within \verb|\boxed{}| in the model’s output and compare it against the reference answer. A binary reward is assigned—1 for exact matches, and 0 otherwise—enabling a clear, scalable and unhackable reward signal for reinforcement learning.

\subsection{Dataset Construction}
\label{sec:data_cur}

To support cognitive transfer, we carefully curate datasets specifically tailored to each training stage, encompassing both language-only and multimodal domains. 

\textbf{Data Collection.} We firstly broadly collect prompt-answer pairs to develop both language and multimodal reasoning skills across mathematical, scientific, and logical domains. For language-only scenarios, we utilize public benchmarks including AIME (up to 2023), MATH~\cite{hendrycksmath2021}, Numina-Math~\cite{numina_cite}, Tulu3 MATH~\cite{tulu3_cite}, and OpenR1-Math-220k~\cite{openr1}, and other open-source datasets. We also synthesize general logical problems via programmatic generation to further enrich reasoning diversity. Multimodal scenarios incorporate datasets covering geometry problem solving (Geometry3k~\citep{geo3k}, GeoQA~\citep{chen2021geoqa}, Geos~\citep{xu2025geosense}), visual discrimination (IconQA~\citep{lu2021iconqa}, Pixmo~\citep{deitke2024molmo}, ChartQA~\citep{masry2022chartqa}), visual puzzles (PuzzleVQA~\citep{chia2024puzzlevqa}, AlgoPuzzleVQA~\citep{ghosal2024language}), STEM (TQA~\citep{kembhavi2017you}, ScienceQA~\citep{lu2022learn}, K12 from \citep{meng2025mm}) and multimodal math (AtomThink~\citep{xiang2024atomthink}, in-house curated math).

\textbf{Data Curation.} To refine data quality, we employ a multi-step curation process. \textit{Firstly}, we employ a pre-trained model to automatically filter out samples with high training loss, which typically indicate noise or excessive complexity. \textit{Secondly}, rule-based and model-assisted methods then identify and remove undesirable patterns~\citep{numina_cite}. \textit{Thirdly}, we apply reweighting to balance coverage, down-weighting overrepresented categories while emphasizing rare but valuable instances. To the end, we distill responses from DeepSeek-R1~\citep{dsr1_cite} to construct approximately \textit{2 million cold-start data}. To ensure the unhackability and stability during RL, we further exclude problems incompatible with our reward functions (e.g., proof-style questions) and apply difficulty-based heuristic filtering, removing both overly trivial and infeasible samples to ensure well-calibrated learning. This leaves around \textit{300k multimodal RL data}. Further details refer to the appendix.

\begin{table}[t]
\centering
\small
\caption{\textbf{Comparison on Language Reasoning and General Benchmarks.} Best results are \textbf{bold} and the second-best are \uline{underlined} for \textit{open-source models}. $^\dagger$ indicates metrics reproduced by ourselves.}
\resizebox{\textwidth}{!}{%
\begin{tabular}{lcccccc}
\toprule
\textbf{Model} & \textbf{AIME 2024} & \textbf{AIME 2025} & \textbf{MATH500} & \textbf{GPQA Diamond} & \textbf{MMLU} & \textbf{MMLU-Pro} \\
\midrule
\textbf{\textit{Open-source Models}}\\
 Qwen2.5-7B~\cite{qwen2p5_cite}&6.7$^\dagger$ & 6.7$^\dagger$&77.6$^\dagger$ & 32.8$^\dagger$&\uline{72.6}$^\dagger$ &\uline{57.5}$^\dagger$\\
 Qwen2.5-VL-7B~\cite{qwen2p5_cite}& 6.7$^\dagger$ & 6.7$^\dagger$ & 67.4$^\dagger$ & 31.8$^\dagger$ & 69.6$^\dagger$ & 51.7$^\dagger$ \\
 Open-Reasoner-Zero-7B~\cite{hu2025open}   & 17.9 & 15.6 & 81.4 & 36.6 & -& -\\
 DeepSeek-R1-Distill-Qwen-7B~\cite{dsr1_cite}& 55.5& 39.2$^\dagger$ & 92.8& 49.1& - & - \\
 QwQ-32B-Preview~\cite{team2025qwq}& 50.0& 33.5& 90.6& \uline{54.5}& -& - \\
 Skywork-R1V-38B~\cite{peng2025skywork} & \textbf{72.0}& -& \uline{94.0}& \textbf{61.6}&-& -\\
 ReVisual-R1~\citep{revisualr1} & 53.3 & \uline{43.3} & 89.2 & 47.5 & - & - \\
 \midrule
\textbf{\textit{Close-source Models}}\\
 Gemini-2.0-Flash~\cite{team2023gemini}& 33.4 & 36.7  & 69.0 & 35.4 & - & - \\
 OpenAI-o1-mini~\cite{openaio1_cite}& 63.6& - & 90.0& 60.0& 85.2&80.3\\
 Claude 3.7 Sonnet~\cite{anthropic2023claude}& 20.0& 13.3 &80.4& 61.1&- &80.0\\
 Doubao-1.5-vision-pro-32k~\cite{DoubaoAI2025}& 26.7 &20.0 & 85.2 &56.1&-&-\\
 \midrule
\rowcolor{AliceBlue} OVR-7B  & \uline{63.5}&  \textbf{52.1}&  \textbf{95.3}&  49.8 &  \textbf{77.2}& \textbf{67.9}\\
\bottomrule
\end{tabular}
}
\label{tab:language_reason}
\end{table}
\section{Experiments}
\label{sec:exp}




In this section, we first elaborate our implementation of \textit{Open-Vision-Reasoner} (OVR). Then, we present superior performance across textual benchmarks (\cref{sec:exp_lang_bmk}) and multimodal scenerios (\cref{sec:exp_multi_bmk}).


\subsection{Implementation Details}
\label{detail}


Our model is based on Qwen2.5-VL-7B~\citep{bai2025qwen2dot5} and employs a two-stage training strategy. In the first stage of cold start, we independently fine-tune the LLM module for 5 epochs with a batch size of 640, a sequence length of 64k, and a learning rate of $2 \times 10^{-4}$ leveraging the default Qwen2.5 configuration~\citep{qwen2p5_cite}. During the subsequent stage of reinforcement learning, following Open-Reasoner-Zero~\citep{hu2025open}, we utilize PPO and configure GAE with $\gamma=1$ and $\lambda=1$ to fully capture long-term dependencies crucial for reasoning tasks, enabling stable training. This RL phase proceeds for 900 iterations, during which we adopt a curriculum for the sequence length: it begins at 24k for the first 300 iterations, increases to 32k through iteration 700, and expands to 48k thereafter, with our latest models continuously undergoing this refinement process. We adhere to strict on-policy updates for the policy model and undertake multiple optimization steps for the critic model. 
Please note that our final model is an \textbf{uniform average of several representative intermediate checkpoints}, ensuring balanced and robust performance across various benchmarks. Additional details can be found in the appendix.

\begin{table}[t]
\centering
\small
\caption{\textbf{Evaluation Results on Visual Reasoning Benchmarks.} Best results are \textbf{bold} and the second-best are \uline{underlined} for \textit{open-source models}. $^\dagger$ Indicates results reproduced by ourselves.}
\resizebox{\textwidth}{!}{%
\fontsize{22}{26}\selectfont
\begin{tabular}{lcccccccccc}
\toprule

\multirow{2}{*}{\textbf{Model}} & \textbf{MathVista} & \textbf{MathVision} & \textbf{MathVerse} & \textbf{DynaMath} & \multicolumn{2}{c}{\textbf{WeMath}} & \textbf{LogicVista} & \textbf{MMMU-Pro} & \multicolumn{2}{c}{\textbf{CharXiv}} \\
& & & vision-only & worst & strict & loose & & & reas. & desc. \\

\midrule
\textbf{\textit{SFT Methods}}\\
LLaVA-OneVision-7B~\cite{llavaov} & 62.6 &17.6 &17.6& 9.0& 17.7 &-&32.0&24.1 &23.6&48.7 \\
InternLM-XComposer2.5~\cite{zhang2024internlm} & 64.0 & 17.8 & 16.2 & 8.2 & 14.1 & - & 34.7 &-&-&- \\
InternVL3-8B~\cite{zhu2025internvl3}&70.5 &28.6 &33.9& 23.0 &37.5&-& 43.6&-&37.6&\textbf{73.6}\\
InternVL2.5-8B~\cite{internvl2p5} & 64.5 & 17.0 & 22.8 & 9.4 & 23.5 & - & 36.0 &34.3 &32.9&\uline{68.6} \\
InternVL2-8B~\cite{chen2024far} & 58.3 & 20.0 & 20.4 & 9.2 & 20.2 & - & 33.6 &29.0 &-&- \\
Qwen2-VL-7B~\cite{qwen2vl} & 61.6 & 19.2 & 25.4 & 11.0 & 22.3 & - & 33.3 &30.5 &34.6&58.0 \\
Qwen2.5-VL-7B~\cite{qwen2p5_cite} & 69.2$^\dagger$ & 25.5$^\dagger$ & 41.1 & 21.8 & 31.2$^\dagger$ & 53.1$^\dagger$ & 47.9 &- &\uline{36.4}$^\dagger$&67.3$^\dagger$ \\
QvQ-72B-Preview~\cite{team2024qvq} & 70.3 &  34.9 & 48.2 & \uline{30.7} & 39.0 & - & \textbf{58.2} &- &-&- \\
Kimi-VL-16B~\cite{team2025kimivl}  & 66.0 &21.8 &34.1&18.0&32.3&-&42.7 & - & -&- \\
\midrule
\textbf{\textit{Close-source Models}}\\
Gemini-2.0-Flash~\cite{team2023gemini} & 70.4 & 47.8& 43.6 &42.1 &47.4&-& 52.3 &-&-&- \\
OpenAI-GPT-4o~\cite{hurst2024gpt} & 59.9 &31.1& 40.6 &34.5 &42.9&-&64.4 & - &-&- \\
Claude 3.7 Sonnet~\cite{anthropic2023claude} & 66.8& 41.3 &52.0 &39.7 &58.2 &-&49.3 &- &-&- \\
GPT-4o mini~\cite{gpt4o}  &55.1$^\dagger$  & 27.3$^\dagger$ & 30.0$^\dagger$ & 31.6$^\dagger$ & 31.4$^\dagger$ &48.8$^\dagger$ & 41.4$^\dagger$ & 37.6$^\dagger$ &34.10$^\dagger$ &74.92$^\dagger$ \\
doubao-1.5-vision-pro-32k~\cite{DoubaoAI2025}& 78.6 & 51.5 & 64.7 & 44.9 & 64.2& - & 65.7\\
\midrule
\textbf{\textit{RL-based Methods}}\\
VLAA-Thinker-Qwen2-7B~\cite{chen2025sft} & 59.6 & 19.8 & 33.9 & 15.2 & 30.5 & - & 36.0 &- &-&- \\
VLAA-Thinker-Qwen2.5-7B~\cite{chen2025sft} & 68.0 & 26.4 & 48.2 & 22.4 & 41.5 & - & 48.5 &- &-&- \\
R1-Onevision-7B~\cite{yang2025r1} & 64.1 & 29.9 & 40.0 &- & - & \uline{61.8} & - &-&-&- \\
OpenVLThinker-7B~\cite{deng2025openvlthinker} & 65.3& 23.0 &38.1& 16.8 &35.2&-& 44.5 & - &-&- \\
MM-Eureka-Qwen-7B~\cite{meng2025mm} & 72.6& 28.1& 45.4& 23.0& 21.8&-& 46.3 &- &-&- \\
MMR1-Math-v0~\cite{mmr1}&69.8& 30.7 &42.8& 17.4& 31.9 &-&46.8&-&-&-\\
ThinkLite-7B-VL~\cite{thinklite}&71.6 &24.6 &42.9 &16.5 &41.8 &-&42.7&-&-&-\\
R1-VL-7B~\cite{zhang2025r1}  & 63.5 & 24.7 & 40.0 & - & - & - & - &- & - &- \\
X-REASONER~\cite{liu2025x} & 69.0& 29.6&-&-& -& -&-& \uline{43.0}&- &- \\
VL-Rethinker-7B~\cite{wang2025vl} &\textbf{73.7} &28.4& 46.4& 17.8 &36.3 &-&42.7 & 41.7 & - & -\\
ReVisual-R1~\citep{revisualr1} & \uline{73.1} & \uline{48.8} & \uline{53.6} & 27.5 & 42.0 &-& 52.3\\
WeThink~\citep{wethink} & 70.9 & 27.2 & 44.7 & 24.4 & \textbf{48.0} &-&53.0 &\\
Skywork-R1V-38B~\cite{peng2025skywork} & 60.6 & 42.1& 40.4 & - & 34.1 & - & 50.6 & - & - &- \\
\midrule
\rowcolor{AliceBlue} \textbf{OVR-7B} & 72.1 & \textbf{51.8}&  \textbf{54.6}&  \textbf{33.5} &  \uline{44.6} & \textbf{64.8} & \uline{54.8} & \textbf{50.2} & \textbf{44.5} & \textbf{73.6}\\
\bottomrule
\end{tabular}
}

\label{tab:math-vl}
\end{table}

\subsection{Enhanced Language Reasoning and General Capabilities}
\label{sec:exp_lang_bmk}
Our model is initially evaluated on a variety of language benchmarks, which cover mathematical reasoning and general problem-solving skills. Specifically, we include \textit{AIME 2024}, \textit{AIME 2025}~\citep{balunovic_srimatharena_2025}, \textit{MATH500}~\citep{hendrycksmath2021}, \textit{GPQA Diamond}~\citep{rein2023gpqa}, \textit{MMLU}~\citep{hendrycksmeasuring}, and \textit{MMLU-Pro}~\citep{mmlu_pro_cite}. 
We compare \textit{Open-Vision-Reasoner (OVR)} with strong LLM baselines, including \textit{Qwen2.5-7B}~\citep{yang2024qwen2}, \textit{DeepSeek-R1-Distill-Qwen-7B}~\citep{dsr1_cite} and \textit{Open-Reasoner-Zero-7B}~\citep{hu2025open}.

OVR demonstrates exceptional language reasoning capabilities. On the challenging AIME 2024 and 2025 benchmarks, it dramatically \textbf{surpasses other 7B open-source models by an average of over 10\%}, achieving performance comparable to leading 32B models. This superiority extends to general reasoning tasks, with significant gains of \textbf{+4.6\%} on MMLU and \textbf{+10.4\%} on MMLU-Pro over parameter-matched competitors. These results highlight the effectiveness of our curated, high-quality cold-start training data.

\subsection{Superior Visual Reasoning Abilities}
\label{sec:exp_multi_bmk}

To evaluate whether the introduced cognitive behavior transfer leads to cross-modal benefits, we further assess the model on a suite of multimodal reasoning benchmarks. These tasks involve image-grounded mathematical reasoning, general multimodal reasoning, and chart understanding. Specifically, we include \textit{MathVista}~\citep{lu2023mathvista}, \textit{MathVision}~\citep{wang2024measuring}, \textit{MathVerse}~\citep{zhang2024mathversedoesmultimodalllm}, \textit{DynaMath}~\citep{zou2024dynamic}, \textit{WeMath}~\citep{qiao2024we}, \textit{LogicVista}~\citep{xiao2024logicvistamultimodalllmlogical}, \textit{MMMU-Pro}~\citep{yue2024mmmuprorobustmultidisciplinemultimodal}, and \textit{CharXiv}~\citep{wang2024charxiv} for evaluation.
We compare our model against strong MLLM baselines, including SFT-based methods such as \textit{LLaVA-OneVision}~\citep{li2024llava} and \textit{Qwen2.5-VL}~\citep{bai2025qwen2dot5}, as well as recent rule-based RL methods like \textit{OpenVLThinker}~\citep{deng2025openvlthinker}, \textit{MM-Eureka}~\citep{meng2025mm} and \textit{ReVisual-R1}~\citep{revisualr1}.


As shown in \cref{tab:math-vl}, our model sets a new breakthrough for 7B models in visual reasoning. 
It is \textbf{the first post-trained Qwen2.5-VL-7B-based model to surpass the 50\% performance on MathVision}, while also achieving state-of-the-art results among 7B models on DynaMath and MathVerse. This strong overall performance is further underscored by a substantial gain on MMMU-Pro (+7.2\% over prior SOTA methods). These results indicate that reasoning capabilities acquired through language training can effectively transfer to multimodal tasks, leading to notable improvements in visual reasoning.

\begin{figure}[t]
    \centering
    \includegraphics[width=0.48\textwidth]{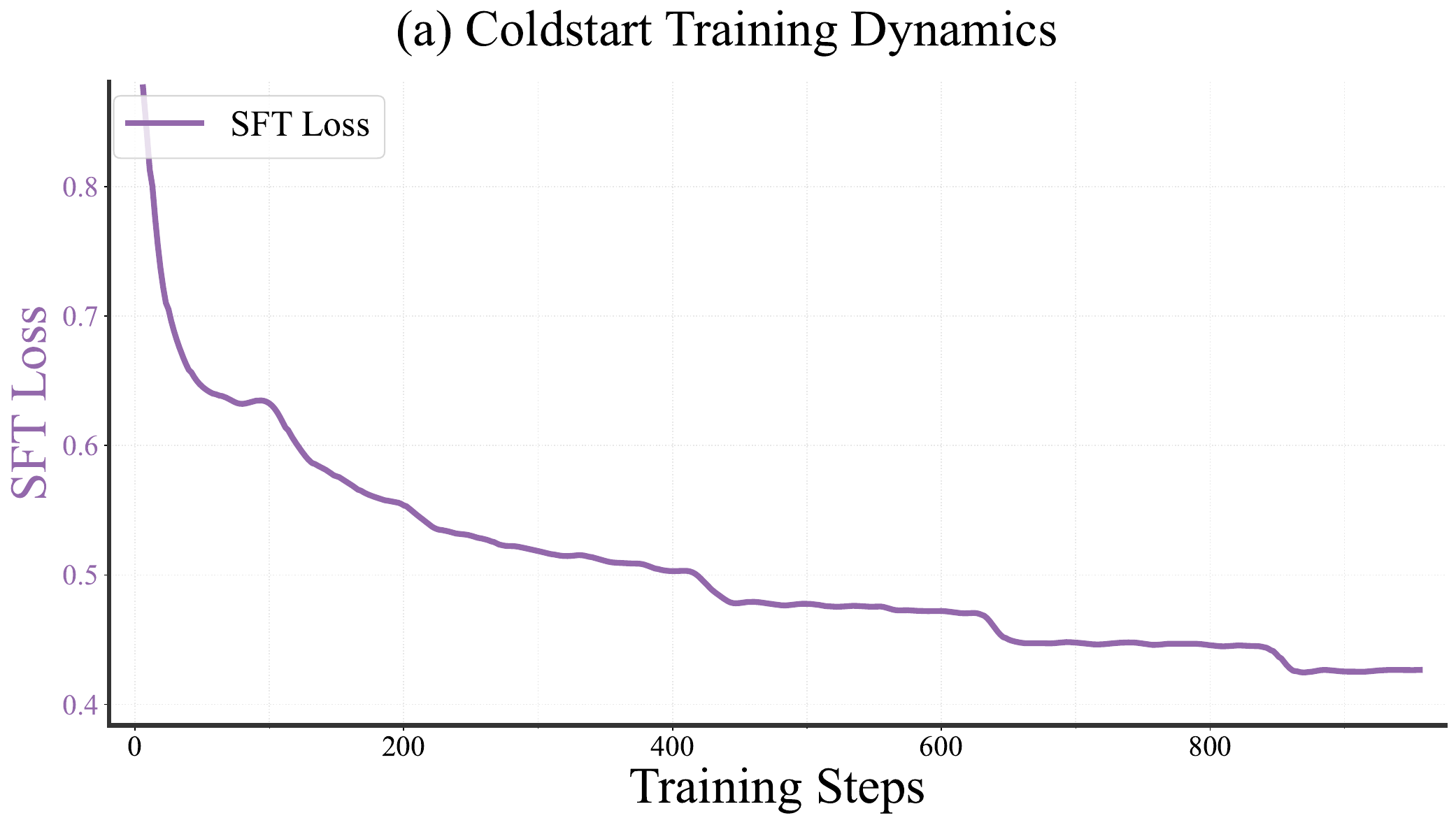}
    \hfill
    \includegraphics[width=0.48\textwidth]{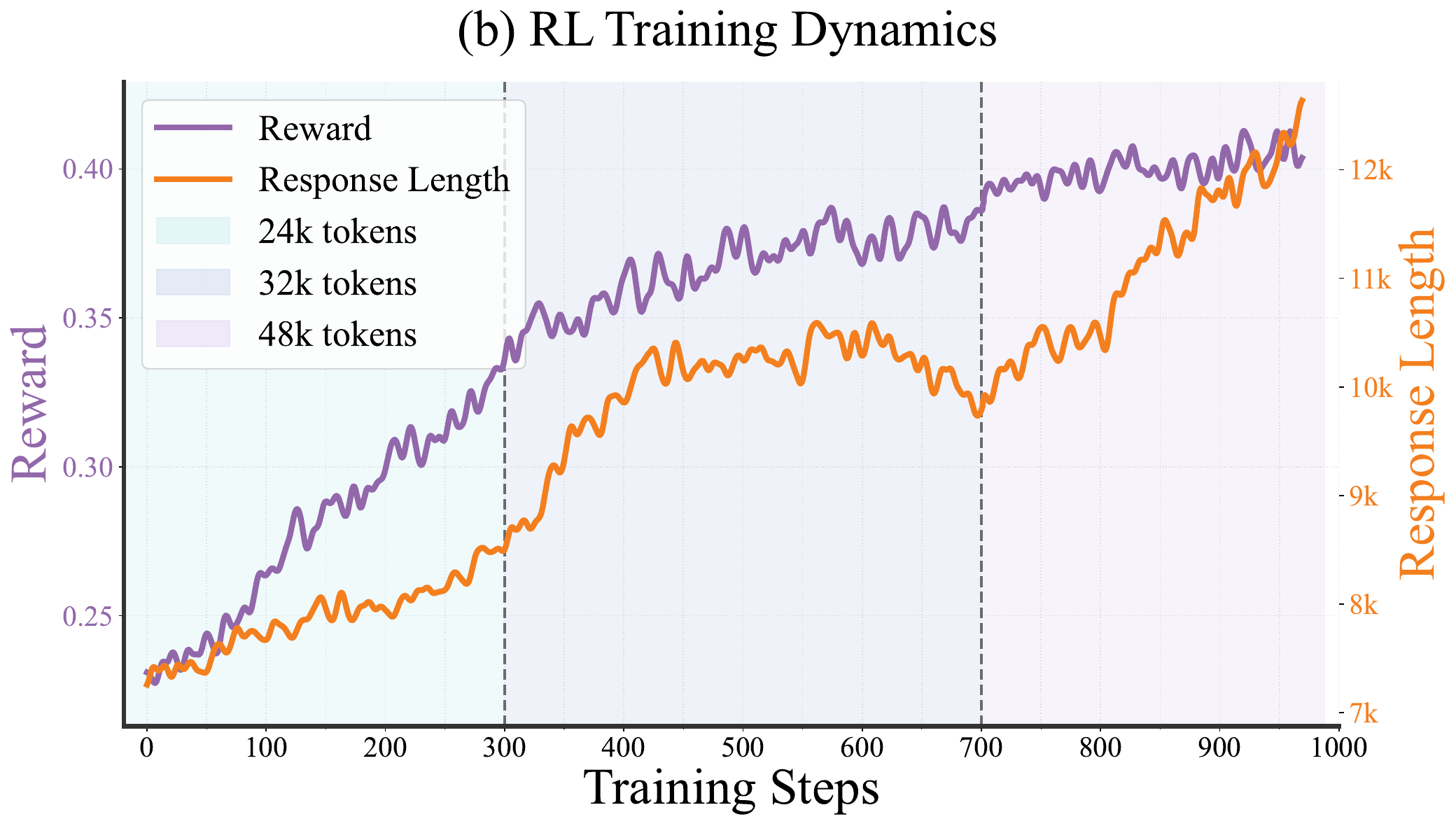}
    \caption{\textbf{Training Dynamics.} (a) The cold-start stage shows a step-wise loss decrease. (b) In the RL stage, reward (purple, left axis) and average response length (orange, right axis) grow steadily, with sharp surges after each sequence length expansion.}
    \label{fig:training_dynamics}
    
\end{figure}

\section{Discussion}

\subsection{Analysis of Training Dynamics}\label{dynamics}

\begin{figure*}[t]
    \centering
    \includegraphics[width=\textwidth]{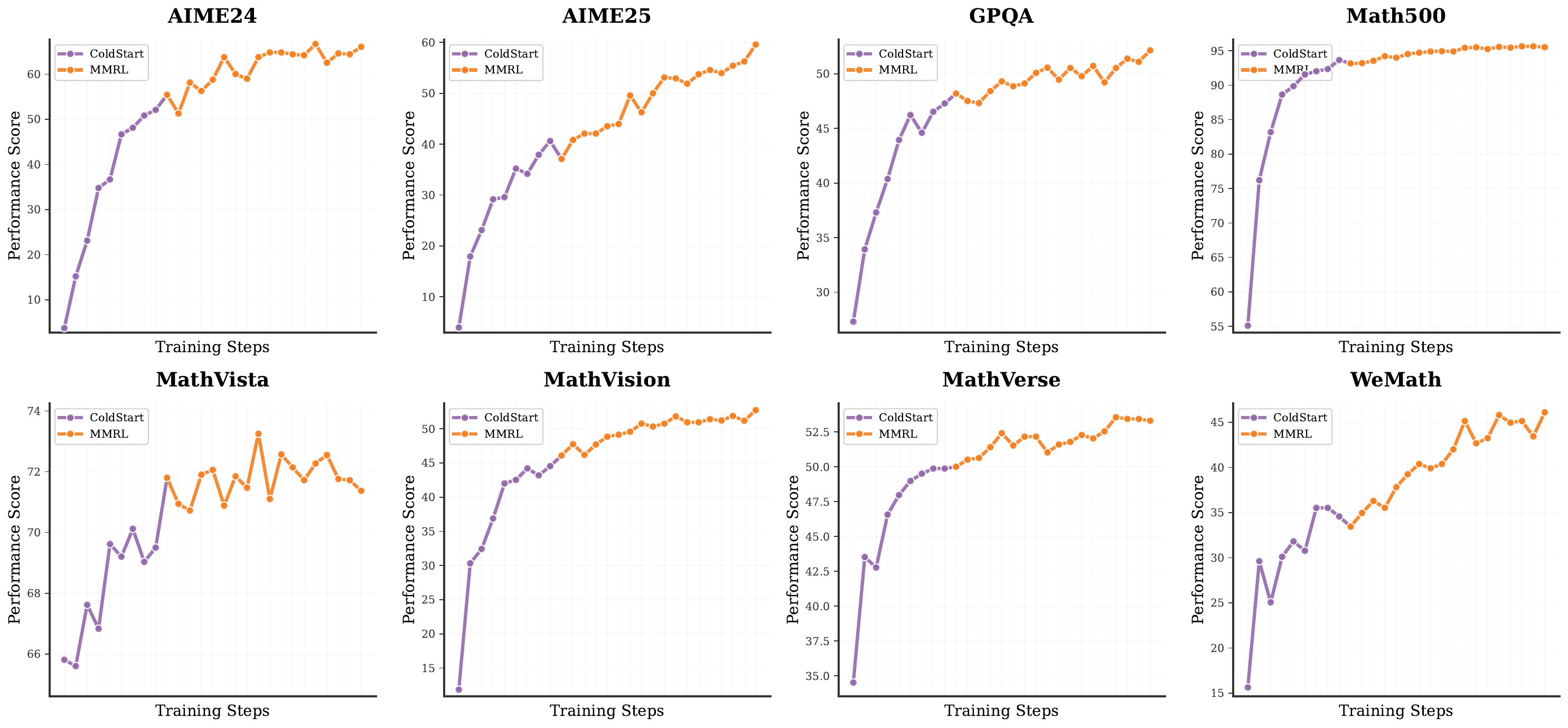}
    
    \caption{\textbf{Performance Evolution on Reasoning Benchmarks.} OVR demonstrates sustained and convergent growth across both linguistic and multi-modal benchmarks throughout the cold start (left) and RL (right) training phases.}
    \label{fig:performance_dynamics}
\end{figure*}

In this section, we present a comprehensive overview of the training dynamics as illustrated in~\cref{fig:training_dynamics}, and provide a detailed analysis of how text and multi-modal reasoning metrics evolve throughout the process as shown in~\cref{fig:performance_dynamics}.

During the initial cold-start phase (\cref{fig:training_dynamics} (a)), the model's loss rapidly descends to below 0.5. Subsequently, across multiple training epochs, the loss exhibits a step-wise, gradual decrease. In parallel, we observe a corresponding surge in performance across all benchmarks (\cref{fig:performance_dynamics}), which first ascend sharply before transitioning to a phase of slower, more incremental improvement toward their peak. A noteworthy observation is that the \textit{aggressive} training strategy detailed in~\cref{detail}—employing a large batch size in concert with a high learning rate—proves to be essential. This approach is critical for breaking the model's inherent constraints, thereby successfully imbuing it with new cognitive paradigms and \textbf{sculpting a more favorable landscape for reinforcement learning}. It is a prerequisite that enables our model, which originates from an instruction-tuned base, to ultimately achieve text performance that is comparable to, or even surpasses models initialized from base~\citep{qwen2p5_cite} or math-specific checkpoints~\citep{qwen2p5math_cite}.

Furthermore, \cref{fig:training_dynamics} (b) reveals how the model's reward and average token length in the RL phase steadily advance from an initial 7k to exceed 12k. Owing to the stability of the training configuration inherited from \textit{Open-Reasoner-Zero}~\citep{hu2025open}, OVR is successfully trained on a diverse corpus of over 20 multi-modal and language-only datasets without encountering any training collapse or performance degradation. Critically, whenever the token length begins to plateau or even decline, we strategically switch to a longer context length, which invariably catalyzes the next wave of rapid reward growth. \cref{fig:performance_dynamics} captures the coincident yet unsurprising convergent growth trajectory shared by all eight reasoning benchmarks, spanning both text and multi-modal domains, as they progressively ascend towards their zenith amidst fluctuations.

\subsection{Multimodal Cognitive Behavior Analysis}
\label{sec:exp_visual_behavior}

Recent studies have highlighted the emergence of cognitive behaviors in MLLMs during visual reasoning tasks—phenomena often dubbed ``visual aha moments''~\citep{chen2025r1v,chen2025sft}. In this work, we move beyond plain observations and systematically investigate how these behaviors are transferred from their linguistic counterparts. Our analysis centers on the four pivotal visual cognitive behaviors introduced in \cref{sec:pre} which are drawn from foundational research on cognitive patterns~\citep{gandhi2025cognitive}. To quantify this process, we employ GPT-4o~\citep{gpt4o} to analyze the emergence of each behavior within the inference traces of our OVR model.

\begin{figure}[t]
    \centering
    \includegraphics[width=0.48\textwidth]{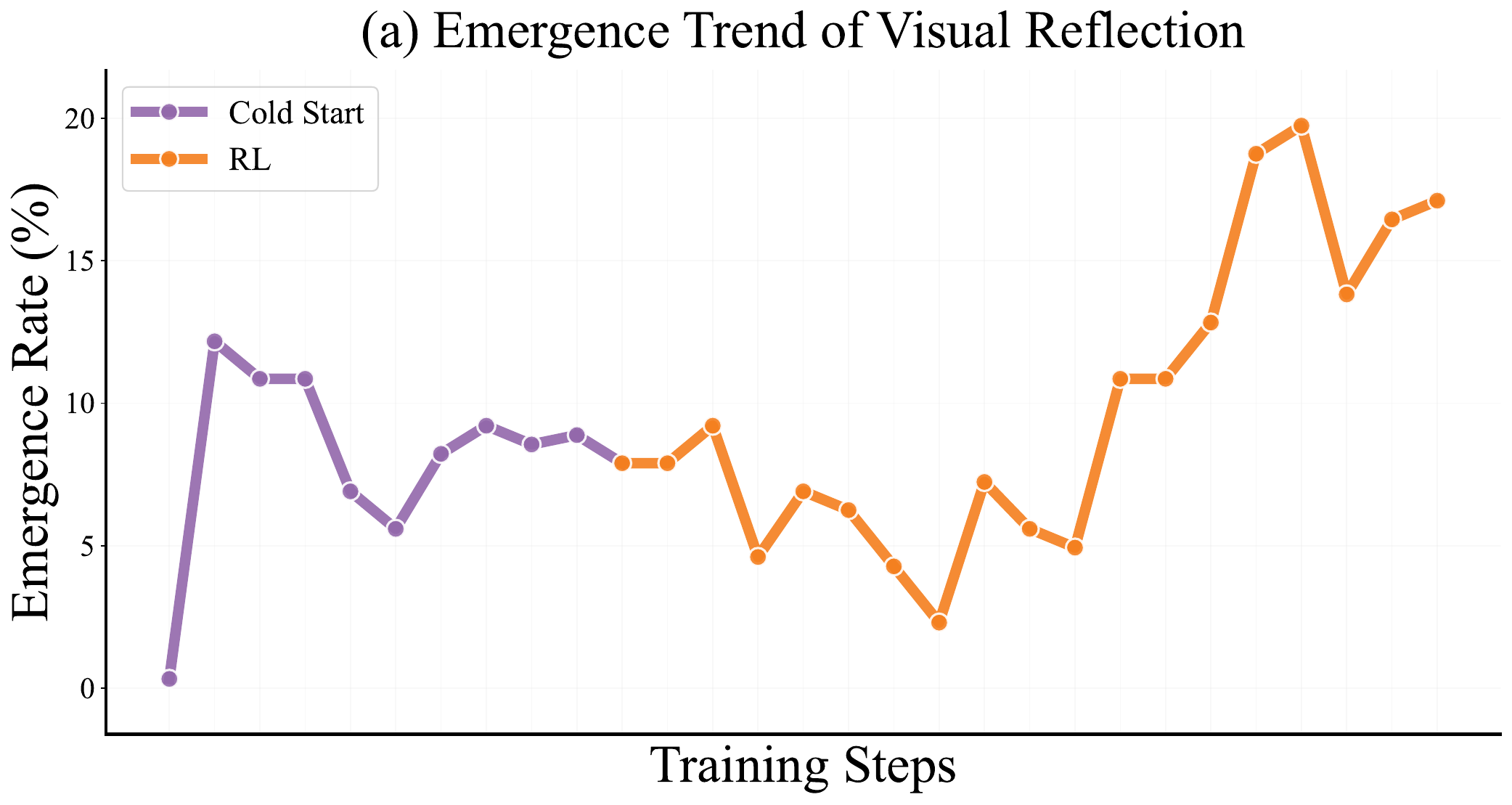}
    \hfill
    \includegraphics[width=0.48\textwidth]{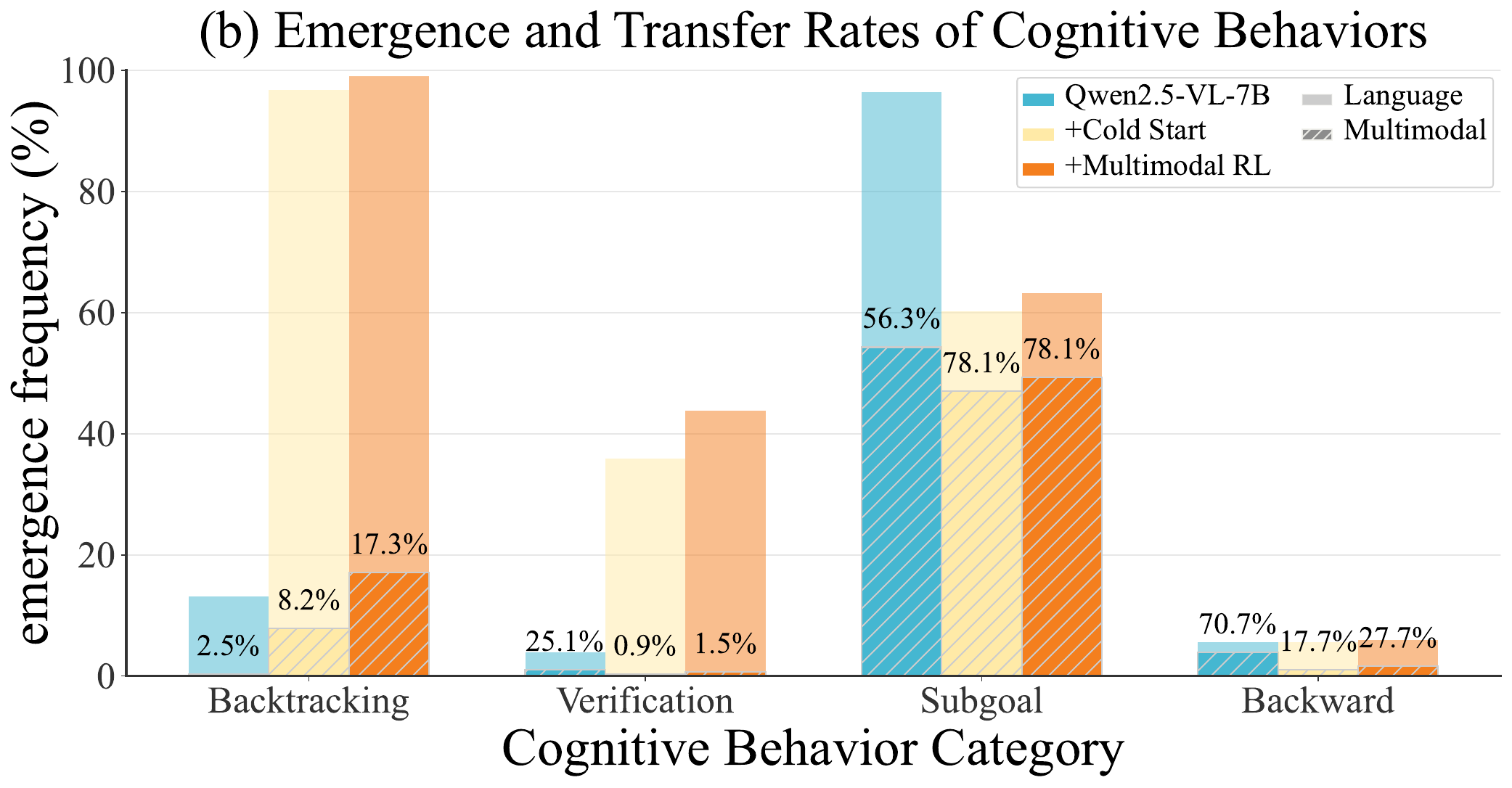}
    \caption{\textbf{Multimodal Cognitive Behavior Analysis.} (a) Emergence of visual reflection across the cold start and RL training steps. (b) Emergence and transfer rates of four visual cognitive behaviors across base models and training stages. Numerical values denote the language-to-vision transfer rates for each behavior.}
    \label{fig:be-sta}
    
\end{figure}

\paragraph{Visual behaviors emerge remarkably early from cold start.}
Following~\cref{dynamics}, we tracked the dynamics of visual reflection, a significant behavior mentioned in previous studys~\citep{wei2025perception,score}, throughout OVR's training. 
As depicted in~\cref{fig:be-sta}, this vision-specific behavior emerges in significant quantities from the \textit{very beginning} of the cold-start phase and fluctuates throughout subsequent training steps.
Strikingly, we observed that even in linguistic problems, DeepSeek-R1's responses~\citep{dsr1_cite} frequently exhibited signs of \textit{mental imagery}~\citep{ms1,ms2} as shown in \cref{fig:r1-lang-case}(a).
The model appeared to construct internal visualizations to support mathematical reasoning, often articulated through phrases such as ``\textit{let me visualize…}'' or ``\textit{let me see the image.}'' 
Once this linguistic scaffolding was introduced into our MLLM, these \textit{mental images} were rapidly grounded in actual visual input, enabling their rapid and effective generalization within OVR.

\paragraph{Cold-start learns broadly, large-scale RL discerns critically.}
We further investigate how cognitive behaviors scale during large-scale RL.
As shown in~\cref{fig:be-sta}(a), after an initial, rapid instillation of patterns during the aggressive cold-start phase, their prevalence is \textit{first suppressed then amplified to unprecedented levels} during multimodal RL.
This counter-intuitive dynamic suggests a clear division of labor: the cold-start phase learns broadly, indiscriminately memorizing all available patterns. In contrast, RL discerns critically, acting as a strategic filter for the crucial tokens~\citep{cheng2025entropy} and scaling up pivotal behaviors. This process of RL—\textbf{discarding the dross to select the essence}—is significant for achieving superior generalization.


\paragraph{Visual transfer of cognitive behaviors is strategic.}

To analyze the transition from linguistic to visual cognition, we track the emergence and transfer rates (detailed in~\cref{btr}) of four core cognitive behaviors across both language and vision modalities. As shown in~\cref{fig:be-sta}(b), the emergence of backtracking and verification steadily increases across training stages, underscoring their growing importance.
Among these, the transfer rate of backtracking shows consistent growth—from 2.5\% to 17.3\%—while verification exhibits near-zero transfer throughout both the cold-start and RL phases.
This indicates that transfer is a \textit{strategic} process, for which we posit two potential explanations:
(1) Backtracking transfers more readily due to DeepSeek-R1’s~\citep{dsr1_cite} inherent ``mental imagination'' capabilities, while verification, lacking a direct linguistic precursor, is more difficult for the MLLM to internalize.
(2) Mirroring how humans naturally and instinctively process visual information~\citep{wei2025perception}, backtracking is a more \textit{fundamental} component of complex visual reasoning, making its amplification a higher priority during the strategic RL phase.
We will investigate these hypotheses in greater depth in our future work.


\subsection{Beyond Behavior: Visual Perception Analysis and Future Work}

Beyond behavioral dynamics, we extend our discussion to a essential capability of MLLMs: \textbf{visual perception} under the cold start plus large-scale RL paradigm. In particular, we investigate two key areas of interest—\textit{perceptual hallucination} and \textit{scaling properties}—through a dedicated study on our OVR model.

\paragraph{Cold start impairs perception, while RL enhances.}
We evaluated both stages of OVR, along with the base model Qwen2.5-VL-7B, on a comprehensive set of multimodal benchmarks targeting visual perception and recognition (MMBench~\citep{mmbench}, BLINK~\citep{blink}, MMStar~\citep{mmstar}, HallusionBench~\citep{hallusionbench}, POPE~\citep{pope}, RealWorldQA~\citep{realworldqa}, MME~\citep{mme}, MMVet~\citep{mmvet}).
As shown in~\cref{tab:model_performance_final}, performance steadily improves across tasks such as MMBench, underscoring the effectiveness of our training paradigm.
The cold-start model shows declines on several tasks, notably increased \textit{hallucinations}~\citep{hall1,hall2}, likely due to token distribution shifts from large-scale linguistic data~\citep{pope}.
However, the regained performance on benchmarks such as MMBench and BLINK demonstrate that long-term multimodal RL can effectively mitigate these issues by \textbf{discerning perceptual capabilities} that are critically for multimodal tasks. 
Looking ahead, degradation from cold start can be mitigated either by incorporating the linguistic data into model pretraining~\citep{mimovl,seed1p5vl}, or by introducing more multimodal supervision during the cold start to establish a stronger visual foundation.

\paragraph{The current unscalability of RL for perception policy.}

\begin{wrapfigure}[11]{r}{0.45\textwidth}
    \vspace{-10pt}
    \centering
    \includegraphics[width=\linewidth]{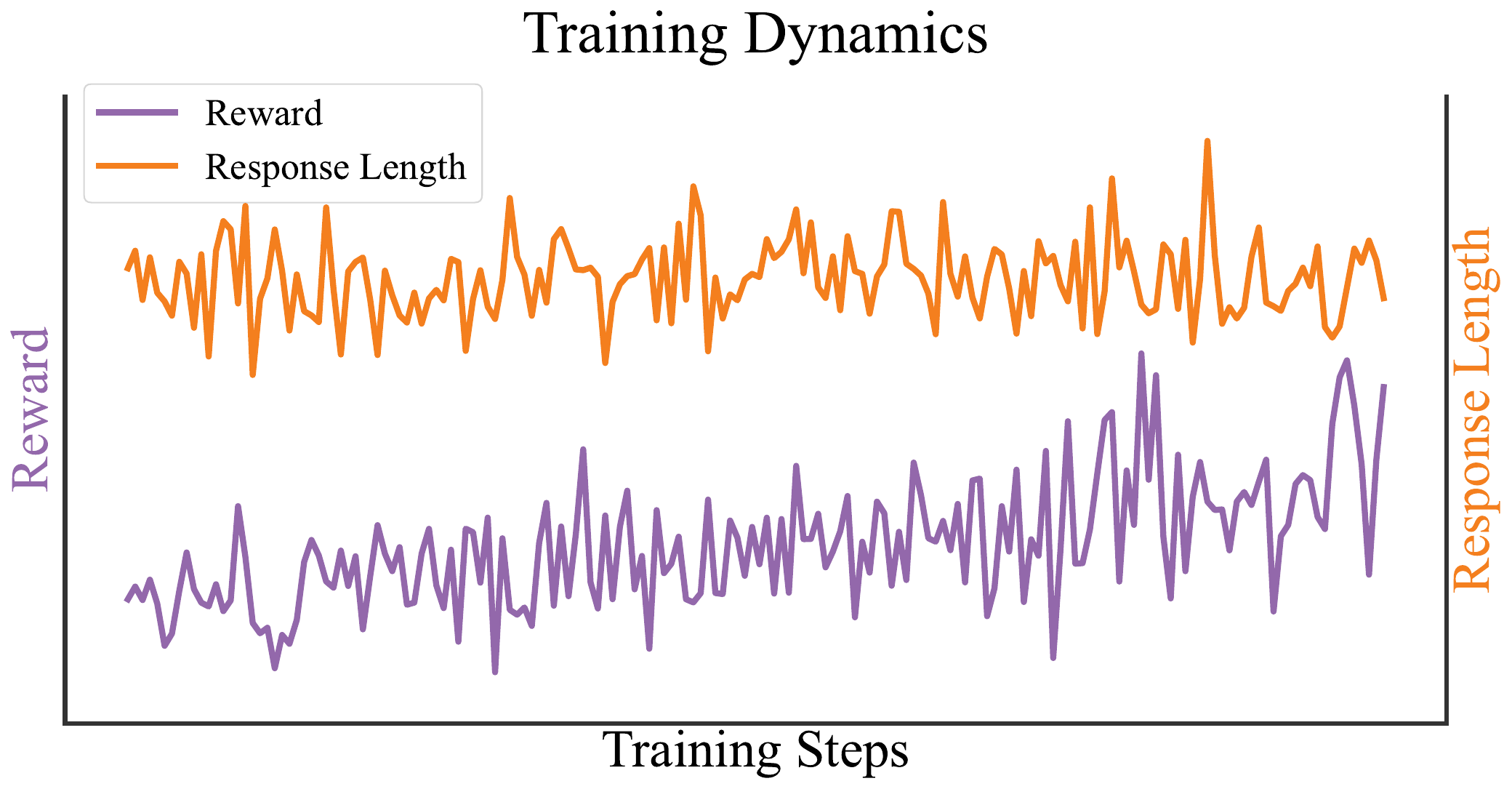}
    \caption{\textbf{Training Dynamics on perception tasks including OCR and counting.}}
    \label{fig:perceptual_training_dynamics}
\end{wrapfigure}

Throughout the multimodal RL, we observed a strong correlation between the reward and the average response length in~\cref{fig:training_dynamics}, which is a finding consistent with prior practices~\citep{dsr1_cite, hu2025open}. 
This reinforces response length as an effective reward proxy, indicative of a \textit{scaling property} tied to reasoning depth and computational resources. 
However, when focusing on specific discriminative perceptual tasks like OCR and counting, we observe a clear divergence. As shown in~\cref{fig:perceptual_training_dynamics}, while the reward can be effectively increased, the average response length remains largely stagnant.

This unscalable training dynamic on such challenging tasks hints at a more fundamental issue: \textit{the absence of certain core visual cognitive behaviors}. Addressing this \textit{fundamental capability gap} is paramount for achieving robust multimodal scaling. Emerging research offers promising avenues, such as multi-turn RL with agentic \textit{tool-use} (\textit{e.g.}, OpenAI-o3~\citep{openaio3_cite}) and the integration of intrinsic imagining through \textit{mental images}~\citep{mentalimagery,spatialmental}. These approaches hold the potential to bridge current limitations and unlock more scalable multimodal reasoning.


\begin{table}[t]
\sisetup{detect-weight}
\centering
\small
\caption{\textbf{Model Performance on Perception-centric Benchmarks.}}
\label{tab:model_performance_final}
\resizebox{\linewidth}{!}{
\begin{tabular}{
  l
  S[table-format=2.1]
  S[table-format=2.1]
  S[table-format=2.1]
  S[table-format=2.1]
  S[table-format=2.1]
  S[table-format=2.1]
  S[table-format=2.1]
  S[table-format=4.1]
  S[table-format=2.1]
}
\toprule
\multirow{2}{*}{\textbf{Model}} & \multicolumn{2}{c}{\textbf{MMBench}} & {\multirow{2}{*}{\textbf{BLINK}}} & {\multirow{2}{*}{\textbf{MMStar}}}& {\multirow{2}{*}{\textbf{HallusionBench}}} & {\multirow{2}{*}{\textbf{POPE}}} & {\multirow{2}{*}{\textbf{RealWorldQA}}} &  {\multirow{2}{*}{\textbf{MME}}} & {\multirow{2}{*}{\textbf{MMVet}}} \\
& {en} & {cn} & & & & & & & \\
\midrule
Qwen2.5-VL & 85.3{$^\dagger$} & \bfseries 84.6{$^\dagger$} & 53.7{$^\dagger$} & 62.1{$^\dagger$} & 49.0{$^\dagger$} & \bfseries 86.1{$^\dagger$}& \bfseries 69.3{$^\dagger$} & \bfseries 1659.7{$^\dagger$}& \bfseries 63.9{$^\dagger$}\\
\rowcolor{AliceBlue!50}
\quad +Cold Start & 86.1& 82.1& 51.5& 62.4& \bfseries 55.0& 82.5& 63.1& 1549.8 & 61.8\\
\rowcolor{AliceBlue!100}
\quad\quad +Multimodal RL & \bfseries 86.6& 84.2& \bfseries 54.1& \bfseries 62.7& 53.6& 83.2& 65.5 & 1559.1& 63.6\\
\bottomrule
\end{tabular}
}
\end{table}

\section{Related Work}

Recent breakthroughs like OpenAI's o1~\citep{openaio1_cite} have highlighted the power of RL in unlocking and scaling reasoning capabilities~\cite{balunovicmatharena,hendrycks2021measuring,rein2024gpqa} within LLMs. DeepSeek-R1-Zero~\citep{dsr1_cite} showed that reasoning capabilities can emerge purely through large-scale RL, leading to complex behaviors like self-verification and reflection. Open-source efforts like Open-Reasoner-Zero~\citep{hu2025open} further demonstrates that even minimalist RL approaches, such as vanilla PPO~\citep{ppo_cite} with GAE~\citep{gae_cite} and simple rule-based rewards, can drive scaling in response length and benchmark performance on open-source models~\citep{qwen2p5_cite,qwen2p5math_cite}.


MLLMs~\citep{bai2023qwen,qwen2vl,bai2025qwen2dot5,dream,AlphaOne25} have rapidly progressed from basic image captioning~\citep{liu2023visual,wei2024vary} to more challenging reasoning tasks~\citep{yue2024mmmu,yu2024merlin,zhao2023chatspot}. Early efforts primarily relied on supervised fine-tuning with Chain-of-Thought (CoT) datasets~\citep{xu2024llava}, while some explored explicit reflection~\citep{wei2025perception} and self-correction~\citep{he2024self} mechanisms to emulate human-like reasoning patterns.
More recently, methods such as PerPO~\citep{zhu2025perpo} and MDPO~\citep{wang2024mdpo} adopt RL-based post-training approaches like DPO~\citep{rafailov2023direct}, where alignment is learned from paired positive/negative responses. These approaches generally follow the RL from Human Feedback (RLHF)\citep{ouyang2022training,yu2024rlhf} or RL from AI Feedback (RLAIF)~\citep{lee2024rlaifvsrlhfscaling} paradigms, where signals from learned reward models or preference labels are utilized for optimization.


Inspired by the success of RLVR~\citep{dsr1_cite} in language models, MLLM research has shifted toward rule-based RL like GRPO~\citep{shao2024deepseekmath} into the multimodal domain. This has led to two major lines of efforts: (1) designing task-specific reward objectives~\citep{yu2025perception,chen2025sft,openr1}, and (2) constructing multimodal “thinking” datasets that embed cognitive behaviors within CoT sequences~\citep{shen2025vlm,huang2025vision,deng2025openvlthinker,meng2025mm}.
Additionally, recent powerful MLLMs adopt a language-only cold start~\citep{revisualr1,mimovl}, using verbal reasoning sequences as a foundation for subsequent multimodal learning. These approaches encourages human-like behaviors~\citep{gandhi2025cognitive} or so-called ``visual aha moments'' in the model responses.

Despite these advances in MLLM, a fine-grained understanding of the underlying reasoning mechanisms remains less explored. In contrast, recent study~\citep{gandhi2025cognitive} centered on LLMs posit that effective reasoning is causally linked to the model's acquisition of certain \textit{cognitive behaviors}, such as verification, backtracking, subgoal setting, and backward chaining. 
The test-time studies have observed that invoking these patterns improve performance~\citep{muennighoff2025s1simpletesttimescaling}.
Entropy-based analysis further reveals that regions associated with cognitive tokens are critical for diverse and high-quality reasoning~\citep{cheng2025entropy}. 
The multimodal work like Long-Perceptual-Thoughts~\citep{liao2025longperceptualthoughts} attempts to explicitly instill such patterns by synthesizing long-form multimodal CoT data.

\section{Conclusion}
\label{sec:con}


In this paper, we propose a two-stage training framework to investigate cognitive behavior in the multimodal domain. By combining a linguistic cold start followed by a large-scale multimodal RL, our approach enables effective cross-modal transfer and scaling of cognitive patterns. Our model \textit{Open Vision Reasoner}, the largest open-source RL practice built on Qwen2.5-VL-7B, demonstrates strong performance across both linguistic and perceptual benchmarks. Beyond performance, we provide a systematic analysis of visual cognitive behaviors, revealing how they emerge and evolve through different training stages. We hope our findings inspire future research on cognitively aligned multimodal agents and open up new possibilities for scaling vision-language reasoning through behavior-centered learning.


\clearpage
\nocite{*}
\bibliographystyle{unsrt}  
\bibliography{main} 

\clearpage
\appendix
\vspace{1em}

The appendix includes extended details on data curation (\cref{sec:data-cur-suppl}), implementation (\cref{sec:imple-detail-suppl}), cognitive behavior evaluation (\cref{sec:cog-detail}), and additional case studies (\cref{sec:more-case}).

\section{Cold-Start Data Curation Details}
\label{sec:data-cur-suppl}

As mentioned in \cref{sec:data_cur}, a critical component of our initial policy development is the curation of the cold-start SFT dataset. This stage serves as the foundation for subsequent learning, particularly in shaping the model’s ability to exhibit structured reasoning and cognitive behavior. To this end, we adopt a multi-stage curation pipeline consisting of data collection, filtering, cleaning, and strategic reweighting.
\paragraph{Data Acquisition.} 
We begin by assembling a broad set of prompt-response pairs that span diverse reasoning domains. These include math, science, and general logical reasoning tasks such as  puzzles, deductive tasks, and constraint satisfaction problems. Our sources include a mix of public datasets as illustrated in~\cref{sec:data_cur}. 
\paragraph{Automated Filtering.}
To improve the signal-to-noise ratio, we filter the collected data using a lightweight pretrained LLM as a proxy for quality estimation. Each instance is passed through this model, and those with abnormally high training loss are flagged as noisy, ambiguous, or misaligned. We further apply rule-based and model-assisted pattern detectors to identify and eliminate undesirable data characteristics. 
\paragraph{Difficulty Stratification.}
We explicitly incorporate samples from AMC, AIME, Olympiads, and AoPS forums to ensure difficulty levels. We then stratify the collected samples based on their source and inherent problem complexity to balance coverage across easy, intermediate, and challenging reasoning scenarios.
\paragraph{Reweighting and Balance.}
To address imbalances across domains and formats, we apply a reweighting scheme based on coverage and reasoning relevance. Over-represented formats are down-weighted, while rare but cognitively rich categories are given higher sampling probabilities. This ensures a more uniform distribution of reasoning challenges and minimizes overfitting to dominant patterns.

\section{More Implementation Details}
\label{sec:imple-detail-suppl}
\paragraph{Model and Optimization Setup}
Our model is based on the Qwen2.5-VL~\citep{bai2025qwen2dot5}. During RL, both the policy and critic networks are initialized from the cold-start model. The value head is initialized from a uniform distribution $\mathcal{U}(-\sqrt{5}, \sqrt{5})$ without bias. The policy and critic networks do not share weights during training. We use the AdamW optimizer with $\beta = [0.9, 0.95]$ and no weight decay. Learning rates are set to $1 \times 10^{-6}$ for the policy and $5 \times 10^{-6}$ for the critic. We use constant learning rates with a linear warm-up of 50 steps, and employ sample packing for improved throughput. No KL regularization or entropy bonus is used, demonstrating that vanilla PPO remains stable under our setup.

\paragraph{PPO Training Dynamics}
Each PPO update is based on 512 unique prompts, each generating 16 sampled responses using temperature and top-p sampling (both set to 1.0). To ensure training stability, we enforce strict on-policy updates for the policy: each prompt generation corresponds to a single optimization step. In contrast, the critic performs 4 optimization steps per PPO update. We apply batch-level advantage normalization to stabilize training further.



\section{Details for Coginitive Behavior Evaluation}
\label{sec:cog-detail}
In this section, we detail definitions of metrics in cognitive behavior analysis (\cref{sec:exp_visual_behavior}) and show the prompt for evaluation.

\subsection{Behavior Transfer Rate }\label{btr}
To quantify how well language-acquired cognitive behaviors generalize to the visual modality, we define the Behavior Transfer Rate (BTR) for each behavior type introduced in~\cref{fig:be-sta}. BTR is calculated as the ratio between the emergence rate of visual behaviors and that of their linguistic counterparts.
Formally, we compute the Cognitive Behavior Emergence Rate in the visual modality ($\text{CBR}_{\text{v}}$) and in the language modality ($\text{CBR}_{\text{l}}$), both evaluated on the multimodal benchmark \textit{MathVision (mini)}. The \textbf{BTR} is then defined as:
\[
\text{BTR} = \frac{\text{CBR}_{\text{v}}}{\text{CBR}_{\text{l}}}
\]
This metric reflects the cross-modal transfer efficiency of cognitive behaviors, with higher values indicating stronger behavioral generalization from language to vision.

\subsection{Evaluation Prompt}
\begin{figure}[htbp]
    \centering
    \includegraphics[width=\textwidth]{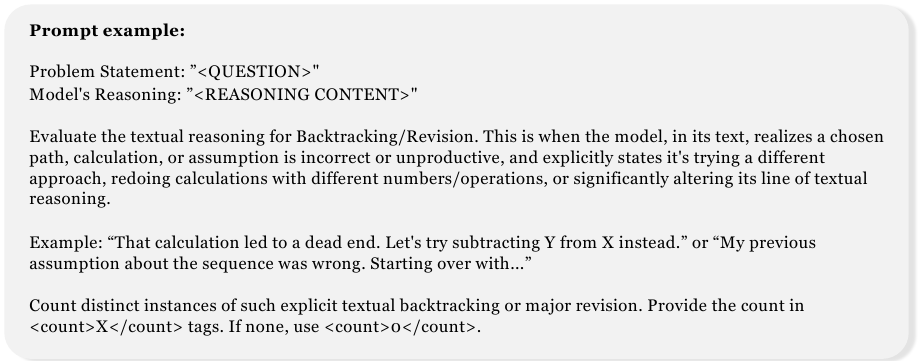} 
    \caption{Prompt design for evaluating \textit{Backtracking}.}
    \label{fig:prompt}
\end{figure}

We design prompts for the LLM-based evaluation. ~\cref{fig:prompt} showcases the prompt template for the cognitive behavior \textit{Backtracking} as an example.

\section{More Case Studies}
\label{sec:more-case}

\begin{figure}[htbp]
    \centering
    \includegraphics[width=0.9\textwidth]{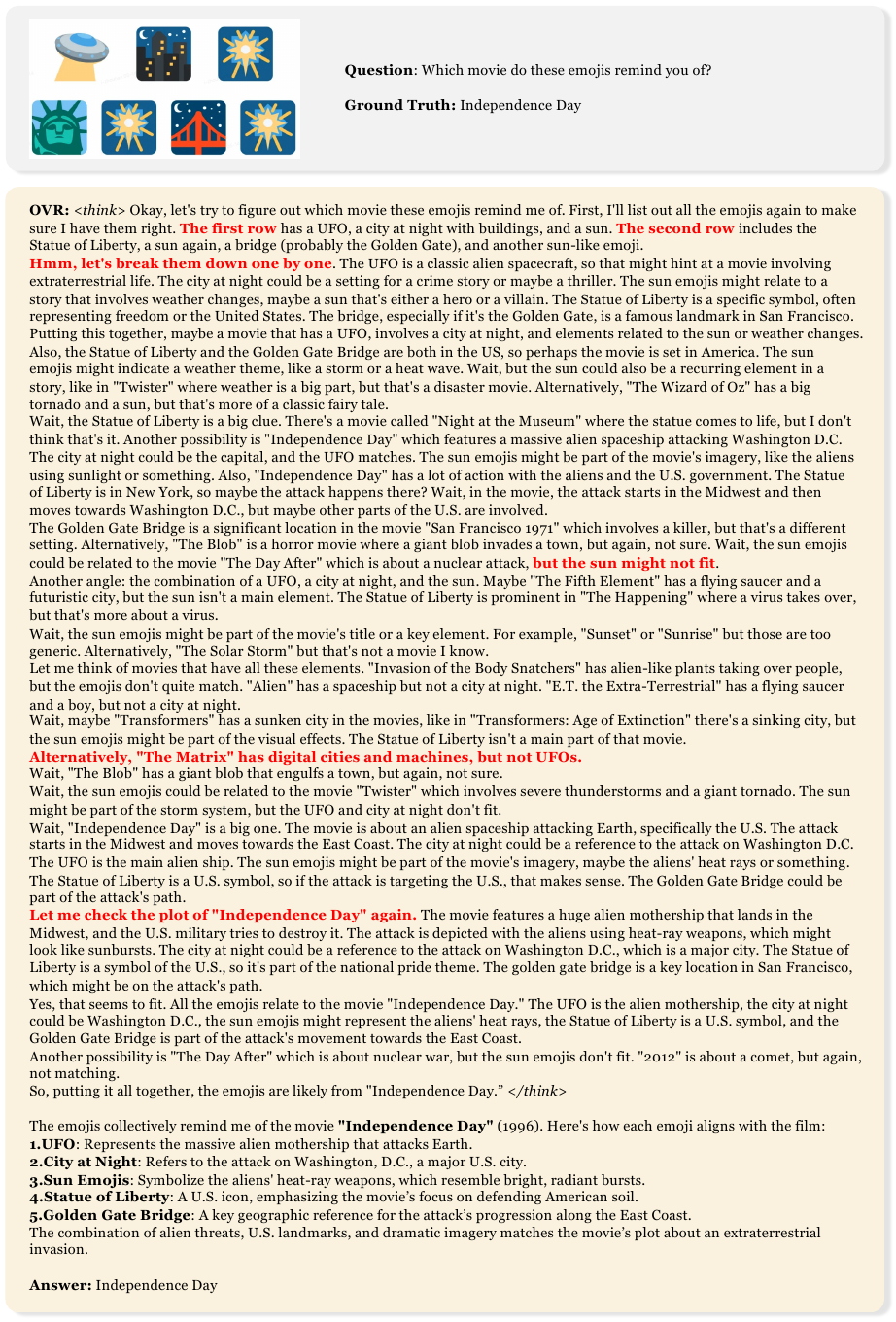} 
    \caption{\textbf{Case Study on Emoji Quiz.} Red highlights some of the cognitive behaviors. The model first performs \textit{visual decomposition} to interpret each emoji individually. It then engages in \textit{subgoal setting} to explore multiple candidate answers, accompanied by \textit{backtracking} to revise earlier hypotheses. The reasoning concludes with a final, justified prediction.
}
    \label{fig:case1}
\end{figure}

We present examples of OVR’s reasoning process during complex multimodal tasks. In \cref{fig:case1}, the model begins by perceiving the image and applying \textit{visual divide-and-conquer} to interpret each emoji individually. It then engages in \textit{subgoal setting}, breaking down the task to explore multiple candidate answers based on each emoji. Throughout the process, the model exhibits \textit{backtracking}, revising earlier hypotheses by \textit{cross-verifying} them with elements in the image. The reasoning process concludes with a final, well-justified prediction.
\begin{figure}[htbp]
    \centering
    \includegraphics[width=0.9\textwidth]{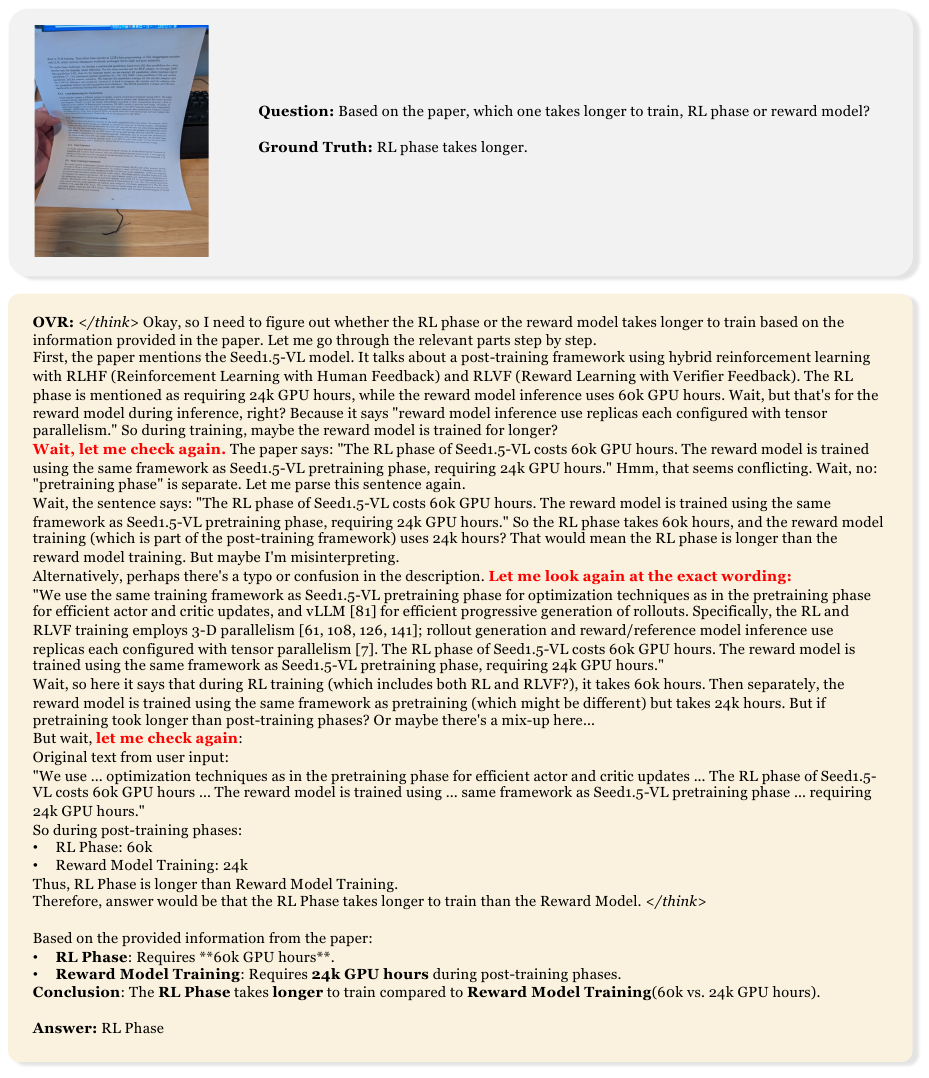} 
    \caption{\textbf{Case Study on OCR-based Document Understanding.} Red highlights some of the cognitive behaviors.}
    \label{fig:case3}
\end{figure}

\begin{figure}[htbp]
    \centering
    \includegraphics[width=0.9\textwidth]{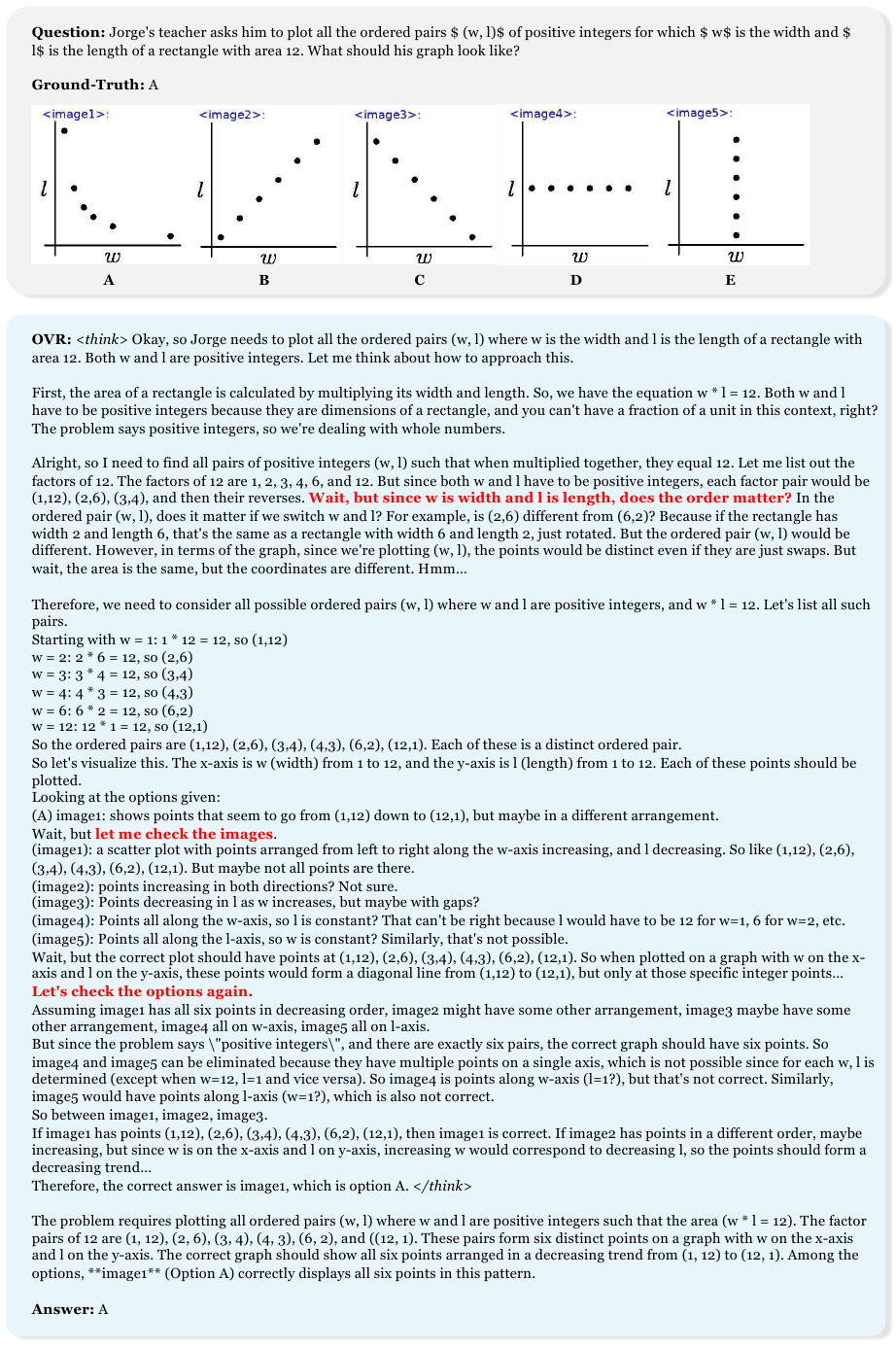} 
    \caption{\textbf{Case Study on Math Reasoning.} Red highlights some of the cognitive behaviors.}
    \label{fig:case3}
\end{figure}

\begin{figure}[htbp]
    \centering
    \includegraphics[width=\textwidth]{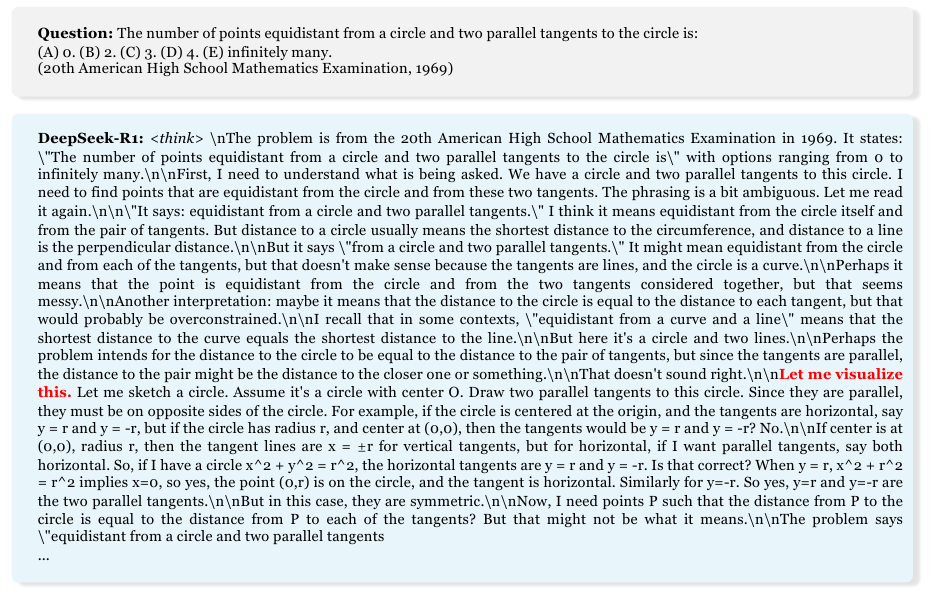} 
    \caption{\textbf{Case Study on DeepSeek-R1.} Red highlights the \textit{mental imagery} mentioned in \cref{sec:exp_visual_behavior}.}
    \label{fig:r1-lang-case}
\end{figure}

\end{document}